\documentclass[journal]{IEEEtran}
\usepackage{cite}

\ifCLASSINFOpdf
   \usepackage[pdftex]{graphicx}
\else
\fi
\usepackage{amsmath,amssymb,amsfonts}
\usepackage{algorithm, algpseudocode}
\begin{document}
%
\title{MILP-based Imitation Learning for HVAC Control }
%
%
%

\author{Huy~Truong~Dinh,~\IEEEmembership{Student Member,~IEEE,}
        and~Daehee~Kim,~\IEEEmembership{Member,~IEEE}
\thanks{This work was supported by the Korea Institute of Energy Technology Evaluation and Planning (KETEP) and the Ministry of Trade, Industry \& Energy (MOTIE) of the Republic of Korea (No. 20184030202130), and this work was supported by the Soonchunhyang University Research Fund.}
\thanks{The authors are with Department of Future Convergence Technology, Soonchunhyang University, Asan 31538, South Korea (e-mail: tdhuy@sch.ac.kr, daeheekim@sch.ac.kr).}
}

%
%

\markboth{Journal of \LaTeX\ Class Files}%
{Shell \MakeLowercase{\textit{et al.}}: Bare Demo of IEEEtran.cls for IEEE Journals}
%

\IEEEpubid{0000--0000/00\$00.00~\copyright~2020 IEEE}


\maketitle

\begin{abstract}
To optimize the operation of a HVAC system with advanced techniques such as artificial neural network, previous studies usually need forecast information in their method. However, the forecast information inevitably contains errors all the time, which degrade the performance of the HVAC operation. Hence, in this study, we propose MILP-based imitation learning method to control a HVAC system without using the forecast information in order to reduce energy cost and maintain thermal comfort at a given level. Our proposed controller is a deep neural network (DNN) trained by using data labeled by a MILP solver with historical data. After training, our controller is used to control the HVAC system with real-time data.  For comparison, we also develop a second method named forecast-based MILP which control the HVAC system using the forecast information. The performance of the two methods is verified by using real outdoor temperatures and real day-ahead prices in Detroit city, Michigan, United States. Numerical results clearly show that the performance of the MILP-based imitation learning is better than that of the forecast-based MILP method in terms of hourly power consumption, daily energy cost, and thermal comfort. Moreover, the difference between results of the MILP-based imitation learning method and optimal results is almost negligible. These optimal results are achieved only by using the MILP solver at the end of a day when we have full information on the weather and prices for the day.
\end{abstract}


%
\IEEEpeerreviewmaketitle

\section{Introduction}

\IEEEPARstart{B}{uilding} sector is not only a major consumer of energy in the electricity market but also plays a key role in reducing of energy consumption in many countries. In the United States, the energy consumption of buildings accounts for $40\%$ of the total energy and $70\%$ of the total electricity \cite{afram2017artificial}. In a building, the energy consumption of heating, ventilation, and air-conditioning (HVAC) systems accounts for $35\%$ of the total energy consumption, and the rest is used for other electrical devices \cite{yoon2020retail}. Therefore, the developments of new techniques to effectively control HVAC systems has always attracted much attention from researchers. Nowadays, with new advances in HVAC systems and new deep learning techniques, many intelligent HVAC control methods have become more feasible and effective.

\begin{figure}[]
\centering
\includegraphics[scale=0.5]{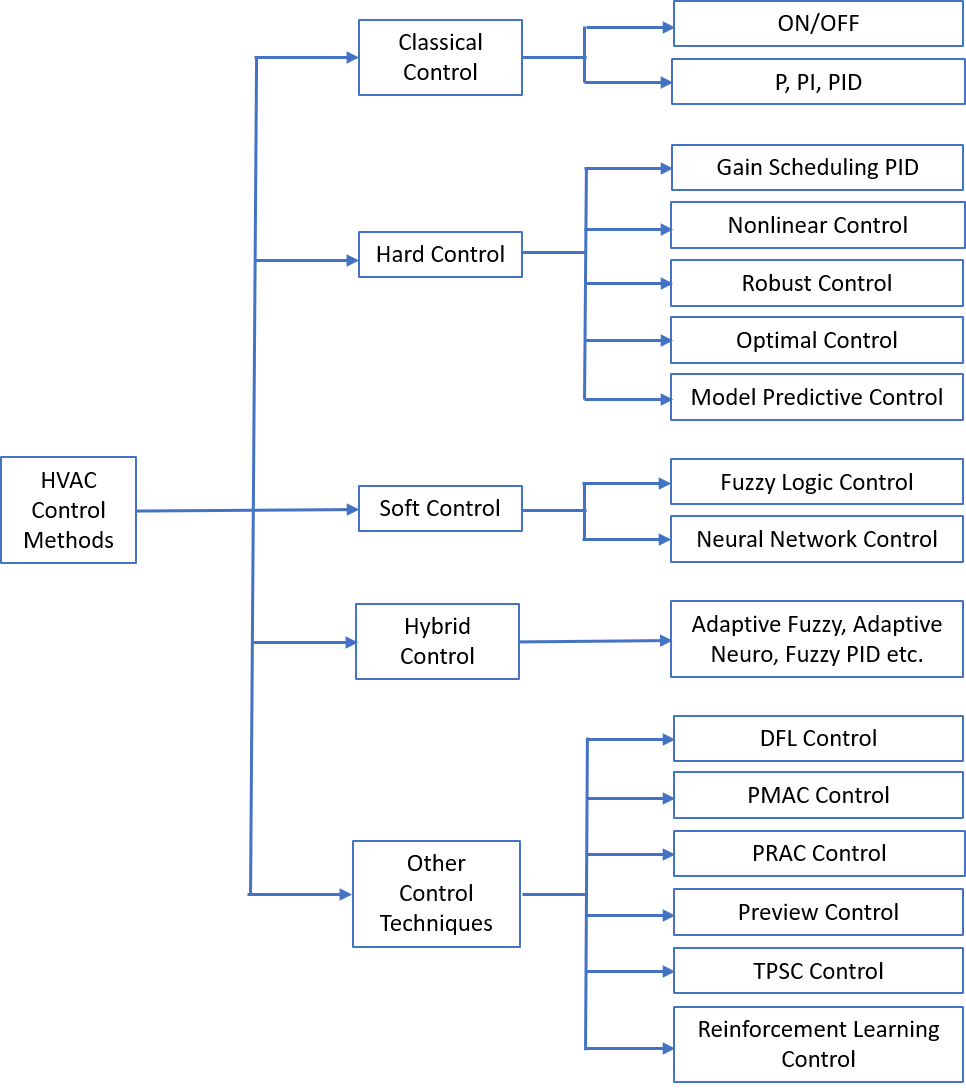}
\caption{Classification of control methods in HVAC systems \cite{afram2014theory}.}
\label{HVAC_systems}
\end{figure}

The HVAC control methods are summarized in Fig. \ref{HVAC_systems} \cite{afram2014theory}. Classical controllers are most commonly used in building energy systems due to their simplicity in design and implementation in practice. In classical methods, the HVAC control is based on logic such as ``if-then-else'' (rule-based controller). For example, upper and lower thresholds are used to turn on/off the HVAC system to guarantee operation within given bounds in the on/off controller. The proportional (P), integral (I), and derivative (D) controllers use a control loop and a feedback mechanism in their operations. In this mechanism, the controllers continuously calculate an error value as the difference between a setup value and a real value and apply a correction based on P, I, and D terms. Classical controllers have been used in many fields such as cooling and heating units \cite{jette1998pi}, room temperature control \cite{jun2011particle}, \cite{bai2008development}, and supply air pressure control \cite{jette1998pi} \cite{pal2008self}. However, they have many disadvantages. A HVAC system controlled by an on/off controller exhibits a large deviation from the optimal value. Parameter auto-tuning in PID controllers is complicated, and the performance of the controllers decreases if the operating conditions differ from the tuning conditions \cite{afram2014theory}. Moreover, auto-tuning parameters cannot be applied to certain applications \cite{salsbury2005survey}.

\IEEEpubidadjcol

Another disadvantage of classical controllers is that they cannot predict future states of the system or take predictive information such as weather and prices into account. Thus, researchers focused on intelligent methods to build HVAC control. Smart controllers can be classified into three categories: hard control, soft control, and hybrid control \cite{afram2014theory}. For example, in hard control, optimal control methods build an optimization model and solve it by using evolutionary algorithms such as genetic algorithm (GA) \cite{mossolly2009optimal}, \cite{yan2008adaptive}. The objectives of the optimization model are generally minimization of energy consumption and maximization of user comfort. Model predictive control (MPC) is one of the most promising methods among hard control methods because of its capability to consider future disturbances and handle multiple constraints and objectives in controller formulation \cite{killian2016ten}. The main idea of MPC is to use a system model to predict the future evolution of the system under disturbances and constraints. At each time step, an optimization problem is solved for the current state and optimal command signals are applied only to operations in this time step. At the next time step, a new optimization problem is solved based on new measurements of the state. However, a serious problem with MPC is the excessive time needed to solve the optimization problem at each step.

To overcome this problem, artificial neural network (ANN) control is introduced as the most promising technique in soft control. To satisfy multiple objectives of users, the optimization problems of HVAC system are usually nonlinear problems, and they are very difficult and time consuming to solve. Hence, an ANN is trained and used to replace nonlinear functions in HVAC control, as in \cite{magnier2010multiobjective}, \cite{javed2016design}. However, the implementation of ANN-based control requires a large training dataset under a wide range of conditions, which may be unavailable for many systems. Moreover, the labels of training data cannot be guaranteed to be best results, thus the ANN's output may not be good enough.

Hybrid controllers employ techniques that are a combination of hard and soft control techniques. These techniques usually include soft control techniques such as ANN at the higher levels and hard control techniques such as MPC controllers at the lower levels \cite{afram2017artificial}, \cite{gouda2006quasi}. Although hybrid control benefits from the advantages of both hard and soft control, it also suffers from the disadvantages of these techniques. For example, training an ANN needs a large of data and labels which sometimes are very difficult to collect.

Currently, reinforcement learning (RL) techniques have been proposed for the control of HVAC systems because of the remarkable success of new RL techniques such as deep Q-Network, deep deterministic policy gradient, and multi-actor-attention-critic \cite{zhang2019building}, \cite{yu2019deep}, \cite{yu2020multi}. The main idea of RL is that the controller uses a DNN to learn from past control actions, and it improves its performance gradually according to rewards. The RL method is generally a model-free method; however, it takes an unacceptably long time to learn and is difficult to implement in practice \cite{wang2008supervisory}.

In this article, we propose a residential HVAC control method for a single zone to reduce energy cost while maintaining a comfortable indoor temperature. Our method is a hybrid control that combines MPC and neural network control. First, our optimization problem is solved by a mix-integer linear programming (MILP) solver with historical data. We then apply imitation learning by using these optimal MILP results to train our controller. After training, at each time slot of a day, this controller can mimic MILP behavior to control the HVAC system with current real-time data. This proposed method adjusts the optimal power input of the HVAC system for only one upcoming time slot by using current information on the weather at this time slot, rather than day-ahead scheduling HVAC system which is based on weather forecast information. We also develop a second method based on forecast information for comparison. To the best of our knowledge, this is the first study that addresses HVAC control using imitation learning. Our main contributions are as follows:
\begin{itemize}
\item We build a nonlinear mathematical model for satisfying both cost reduction and thermal comfort in a single zone equipped with a HVAC system as the residential heating system during a day. Our nonlinear model is also successfully transformed to a MILP model which can be easily solved and for which global optimal results are guaranteed.
\item On the basis of these formulas, we propose two methods for controlling HVAC systems based on a MILP solver: MILP-based imitation learning and forecast-based MILP. In the first method, a DNN with two hidden layers successfully mimics the MILP solver to control the HVAC system effectively.
\item We try to evaluate the two methods by performing extensive simulations with real data. Simulation results show that the performance of the MILP-based imitation learning method is better than that of the second method. Moreover, the hourly power inputs of an HVAC system controlled by MILP-based imitation learning are very similar to the optimal results, which can be calculated by the MILP solver only at the end of the day when we have full information on the outdoor temperature of that day.
\end{itemize}

The rest of this paper is organized as follows. Section II reviews the works related to our methods. In Section III, our problem is formulated. A detailed linear transformation and two proposed MILP-based methods are described in Section IV. In Section V, detailed simulations and numerical results are provided. Finally, Section VI outlines the conclusions and potential future works.

%
%
%
%

\section{Related works}
Because our approaches combine DNN with a MILP solver, we review existing ANN studies and mathematical models for HVAC control in this section.

In \cite{kelman2011bilinear}, HVAC control in a typical commercial building is studied. Authors propose an MPC approach whose objectives are to minimize energy use while satisfying thermal comfort. Sequential quadratic programming (SQP) is applied to solve their nonlinear optimization problem. The computational time of SQP is less than one minute, faster than timescales of a HVAC system. However, the SQP algorithm cannot guarantee global optimal results for nonlinear optimization. Moreover, energy cost is not considered in their study. 

Authors in \cite{magnier2010multiobjective} propose an ANN-based approach for multi-objective (MO) optimization of a residential building in presence of a HVAC system. Two objectives considered are thermal comfort and energy consumption, and MO evolutionary algorithms (MOEAs) are used as the optimization engines. To overcome extremely the time-consuming problem of solving the MO optimization problem, an ANN is used and trained by an MO genetic algorithm (NSGA-II). After that, the ANN is implemented to enable fast evaluations. The simulation results show that the performance of the ANN is very good in energy consumption but generally underestimates in thermal comfort. Moreover, in theory, MOEAs cannot guarantee that global optimal results will be found, and thus the labels of the training ANN may not be good enough.

In \cite{javed2016design} and \cite{javed2015experimental}, authors study a thermal comfort controller based on a random neural network (NN) in HVAC control. In their study, the predicted mean vote of Fanger model is adopted to evaluate the indoor thermal comfort. A hybrid particle swarm optimization (PSO)-SQP algorithm is used to train the random NN. The simulation results show that the ANN-based controller is computationally less expensive in run time and maintains indoor thermal comfort in a desired range. However, the price and energy cost are not considered in their study. Moreover, in theory, global optimal results cannot be guaranteed to be found in the PSO algorithm, and thus the labels of the training random NN may not be good enough.

In terms of the electricity price and energy cost, authors in \cite{lee2015optimal} propose an ANN-based MPC method to control an HVAC system in a zone for the next 24 hours to minimize energy cost and ensure thermal comfort. Their objective functions are formulated as a mixed-integer nonlinear programming (MINLP) problem which is difficult to solve. Nonlinear autoregressive neural networks (NARNETs) are used to simplify MINLP and model the thermal behavior of the zone and the energy cost of a day. The simulation results show that their approach can reduce the energy cost by $14.25\%$ compared with the base case and achieve a comfortable thermal range. However, the authors cannot guarantee global optimal results because a simplified MINLP was used.

In \cite{reynolds2018zone}, authors propose a zone-level, heating set point scheduler for the next 24 hours. In their work, objectives are to minimize energy consumption in a day while maintaining thermal comfort within the multi-zone building. At each time step, an ANN is used to predict the indoor temperature and energy consumption at the next time step by using forecast information regarding the weather, occupancy, and temperature set points. A GA then uses the ANN as an evaluation engine to calculate the energy consumption over 24 hours to find the best heating set point schedules for each zone. With time-of-use tariff, their scheduler can shift the load and reduce the energy cost by $27\%$ compared with the fixed schedule.

Similar to \cite{reynolds2018zone}, in \cite{kim2020supervised}, authors propose two multi-zone HVAC schedulers for the next day based on forecast information such as outdoor temperature and solar radiation. In the first method, a NARNET-based ANN is used to simplify the optimization problem and estimate the zone temperature based on forecast information. Its output is used to calculate the optimal power inputs of the HVAC for 24 time slots of the next day based on a MILP solver. In the second method, a DNN is used to replace the MILP solver to reduce computation time. Simulation results show that their methods can keep the indoor temperature in a comfortable range and shift the load to reduce the energy cost in a day.

However, in previous studies, these schedulers used forecast information to control the HVAC system. Hence, these schedulers depend on the accuracy of the forecast information. Forecast information always contains errors, and these errors are larger in winter with bad weather \cite{martner1999five}. These errors cause schedulers control the HVAC system incorrectly.

To avoid using unstable forecast information and to guarantee global optimal results in training the DNN, we propose MILP-based imitation learning for HVAC control. Our DNN is trained by global optimal results created by a MILP solver and our method controls the HVAC system at each hour in a day based on current real-time information.

\section{Problem Formulation}

In this section, we build mathematical formulas of the HVAC model and our objective function during a day from 0 A.M. to 12 P.M. We divide a day into $T=24$ time slots, and the duration of each time slot is $\Delta t=1$ hours. In this study, a HVAC system with an inverter is  considered. This enables us to adjust the input power of the HVAC system continuously to achieve the desired thermal comfort \cite{7890446}. 

\subsection{Energy cost of HVAC system}

Suppose $p_t$ is the input power of the HVAC system during time slot $t$, then the energy cost of the HVAC system in a day, $C_{day}$, is calculated as follows:

\begin{equation}
C_{day} = \displaystyle \sum_{t=1}^{T} p_t \cdot \Delta t \cdot price_t = \displaystyle \sum_{t=1}^{T} p_t \cdot price_t \hspace{0.5cm} \forall t
\end{equation}
where $price_t$ is the  price of the main grid at time slot $t$. We call $P_{max}$ the rating power of the HVAC system; thus, the input power $p_t$ must satisfy the following constraint.

\begin{equation}
0 \leq p_t \leq P_{max} \hspace{0.5cm} \forall t
\label{constraint_power}
\end{equation}

\subsection{Thermal Comfort}

In general, HVAC systems have two operational modes: heating mode and cooling mode. In this study, we primarily focus on heating mode in the winter as \cite{yu2017online}. According to \cite{yu2019deep}, \cite{yu2017online}, \cite{pilloni2016smart}, \cite{li2020real}, the indoor air temperature adjusted by the HVAC system in heating mode can be obtained as follows:

\begin{equation}
T_{t+1}^{in} = \varepsilon \cdot T_t^{in} + (1- \varepsilon) \cdot (T_t^{out} + \frac{\eta \cdot p_t}{A})\hspace{0.5cm} \forall t
\label{formula_input_temperature}
\end{equation}
where $T_t^{in}$ and $T_t^{out}$ refer to the indoor temperature and outdoor temperature of home at time slot $t$. $p_t$ is the input power of the HVAC system at time slot $t$. $\varepsilon$, $\eta$, and $A$ are constants that describe the specification of the home environment and HVAC device. $\varepsilon$ is the system inertia, $\eta$ is the thermal conversion efficiency, and $A$ is the overall thermal conductivity.

There are many factors that affect thermal comfort inside a smart home such as air temperature, radiant temperature, humidity, and air speed. Thermal comfort also depends on personal factors (e.g., clothing, personal activity, and physical condition). In this study, similar to \cite{yu2019deep}, \cite{li2020real}, the indoor air temperature is considered only as a representation of thermal comfort. The HVAC system is used to adjust the indoor temperature $T_t^{in}$ at time slot $t$ to satisfy the following comfortable thermal range.
\begin{equation}
T_{min} \leq T_t^{in} \leq T_{max}\hspace{0.5cm} \forall t
\label{constraint_input_temperature}
\end{equation}
where $T_{min}$ and $T_{max}$ are the minimum and maximum temperatures of the comfortable range set up by residents. It is worth noting that the range of $p_t$ is $[0, P_{max}]$. When $T_t$ is higher than $T_{max}$, $p_t$ should be zero to prevent further temperature deviation.

To be able to measure the user's thermal comfort at each time slot, we introduce a function of thermal discomfort, $TDC(t)$, which can be calculated as
\begin{equation}
TDC(t) = 
   \begin{cases}
   T_{min} - T_t^{in} & \quad  T_t^{in} \leq T_{min}. \\
   0 & \quad T_{min} \leq T_t^{in} \leq T_{max}. \\
   T_t^{in} - T_{max}  & \quad T_{max} \leq T_t^{in}.
   \end{cases}
\label{formula_discomfort}
\end{equation}

\begin{figure}[]
\centering
\includegraphics[scale=0.5]{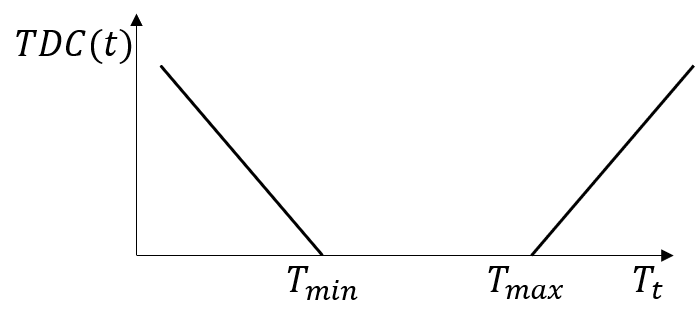}
\caption{Definition of thermal discomfort.}
\label{thermal_discomfort}
\end{figure}

Fig \ref{thermal_discomfort} shows the distribution function $TDC(t)$ where thermal discomfort increases if $T_{t}^{in}$ is outside the comfortable range $[T_{min}, T_{max}]$.

\subsection{Objective Function}
The main objective of HVAC control is to not only reduce the energy cost but also maintain the indoor temperature in a comfortable range that provides thermal comfort. Therefore, our multi-objective (MO) function can be formulated as follows:
\begin{align*}
min(MO) &= min \Big(C_{day} + \alpha \cdot \displaystyle \sum_{t=1}^{T} TDC(t)\Big) \nonumber \\
&= min \bigg(\displaystyle \sum_{t=1}^{T} \Big(p_t \cdot price_t + \alpha \cdot TDC(t) \Big) \bigg)\\
&s.t. \hspace{0.2cm} (\ref{constraint_power}) - (\ref{formula_discomfort}) \nonumber 
\label{MOF}
\end{align*}
where $\alpha$ is a balance parameter for achieving the trade-off between the energy cost and thermal discomfort that is set by the residents.

\section{two MILP-based methods}

Because of the definition of function $TDC(t)$ in (\ref{formula_discomfort}), the above MO function is a nonlinear function and it is difficult to solve. Hence, in this section, we describe a way to convert our MO function to a linear function, and two HVAC control methods based on the mixed-integer linear programming (MILP) solver are proposed.

\subsection{Linear Transformation}
To transform our MO function to a linear function, function $TDC(t)$ is modeled by using binary linear variables and linear constraints. We introduce six new variables that are dependent on the $T_t^{in}$ variable at each time slot $t$ and define them as follows:

\begin{equation}
w_1[t] = 
\begin{cases}
1 & \text{if } T_t^{in} \leq T_{min}  \text { and } w_2[t] \neq 1 \text{ and } w_3[t] \neq 1 \\
0 & otherwise \\
\end{cases}
\label{w_1}
\end{equation}
\begin{equation}
x_1[t] = 
\begin{cases}
T_t^{in} & \text{if } T_t^{in} \leq T_{min} \text { and } x_2[t] \neq T_t^{in} \text{ and } x_3[t] \neq T_t^{in}\\
0 & otherwise \\
\end{cases}
\label{x_1}
\end{equation}
\begin{equation}
w_2[t] = 
\begin{cases}
1 &\text{if } T_{min} \leq T_t^{in} \leq T_{max} \text { and } w_1[t] \neq 1 \text{ and } w_3[t] \neq 1\\
0 & otherwise \\
\end{cases}
\label{w_2}
\end{equation}
\begin{equation}
x_2[t] = 
\begin{cases}
T_t^{in} &\text{if } T_{min} \leq T_t^{in} \leq T_{max} \text { and } x_1[t] \neq T_t^{in} \text{ and } x_3[t] \neq T_t^{in}\\
0 & otherwise \\
\end{cases}
\label{x_2}
\end{equation}
\begin{equation}
w_3[t] = 
\begin{cases}
1 & \text{if }  T_{max} \leq T_t^{in} \text { and } w_1[t] \neq 1 \text{ and } w_2[t] \neq 1\\
0 & otherwise \\
\end{cases}
\end{equation}
\label{w_3}
\begin{equation}
x_3[t] = 
\begin{cases}
T_t^{in} & \text{if } T_{max} \leq T_t^{in} \text { and } x_1[t] \neq T_t^{in} \text{ and } x_2[t] \neq T_t^{in}\\
0 & otherwise \\
\end{cases}
\label{x_3}
\end{equation}
where $w_1[t]$, $w_2[t]$, and $w_3[t]$ are binary variables and $x_1[t]$, $x_2[t]$, and $x_3[t]$ are float variables. Here are the constraints that are equivalent to the definitions of the six variables.

\begin{equation}
w_1[t] + w_2[t] + w_3[t] = 1
\label{w_sum}
\end{equation}
\begin{equation}
x_1[t] + x_2[t] + x_3[t] = T_t^{in}
\label{x_sum}
\end{equation}
\begin{equation}
\mu \cdot w_1[t] \leq x_1[t] \leq T_{min} \cdot w_1[t]
\label{cons_x_1}
\end{equation}
\begin{equation}
T_{min} \cdot w_2[t] \leq x_2[t] \leq T_{max} \cdot w_2[t]
\label{cons_x_2}
\end{equation}
\begin{equation}
T_{max} \cdot w_3[t] \leq x_3[t] \leq M \cdot w_3[t]
\label{cons_x_3}
\end{equation}
where $\mu$ and $M$ are a very small and a very large value that $T_t^{in}$ never reaches. For example, $\mu = -1000$ and $M = 1000$. Applying the definitions from (\ref{w_1}) to (\ref{x_3}), function $TDC(t)$ turns into
\begin{equation}
TDC(t) = T_{min} \cdot w_1[t] - x_1[t] + x_3[t] - T_{max} \cdot w_3[t].
\end{equation}

Finally, we have a new MO function as follows.
\begin{align}
min(MO) &= min \bigg(\displaystyle \sum_{t=1}^{T} \Big(p_t \cdot price_t + \alpha \cdot TDC(t) \Big) \bigg) \nonumber \\
&= min \bigg(\displaystyle \sum_{t=1}^{T} \Big(p_t \cdot price_t + \nonumber \\
&\alpha \cdot \big(T_{min} \cdot w_1[t] - x_1[t] + x_3[t] - T_{max} \cdot w_3[t] \big) \Big) \bigg) 
\label{MOF_2}
\end{align}
\begin{equation}
s.t. \hspace{0.2cm} (\ref{constraint_power}),(\ref{formula_input_temperature}), (\ref{constraint_input_temperature}),(\ref{w_sum}) - (\ref{cons_x_3}) \nonumber \\
\end{equation}

\begin{figure*}[ht]
\centering
\includegraphics[scale=0.5]{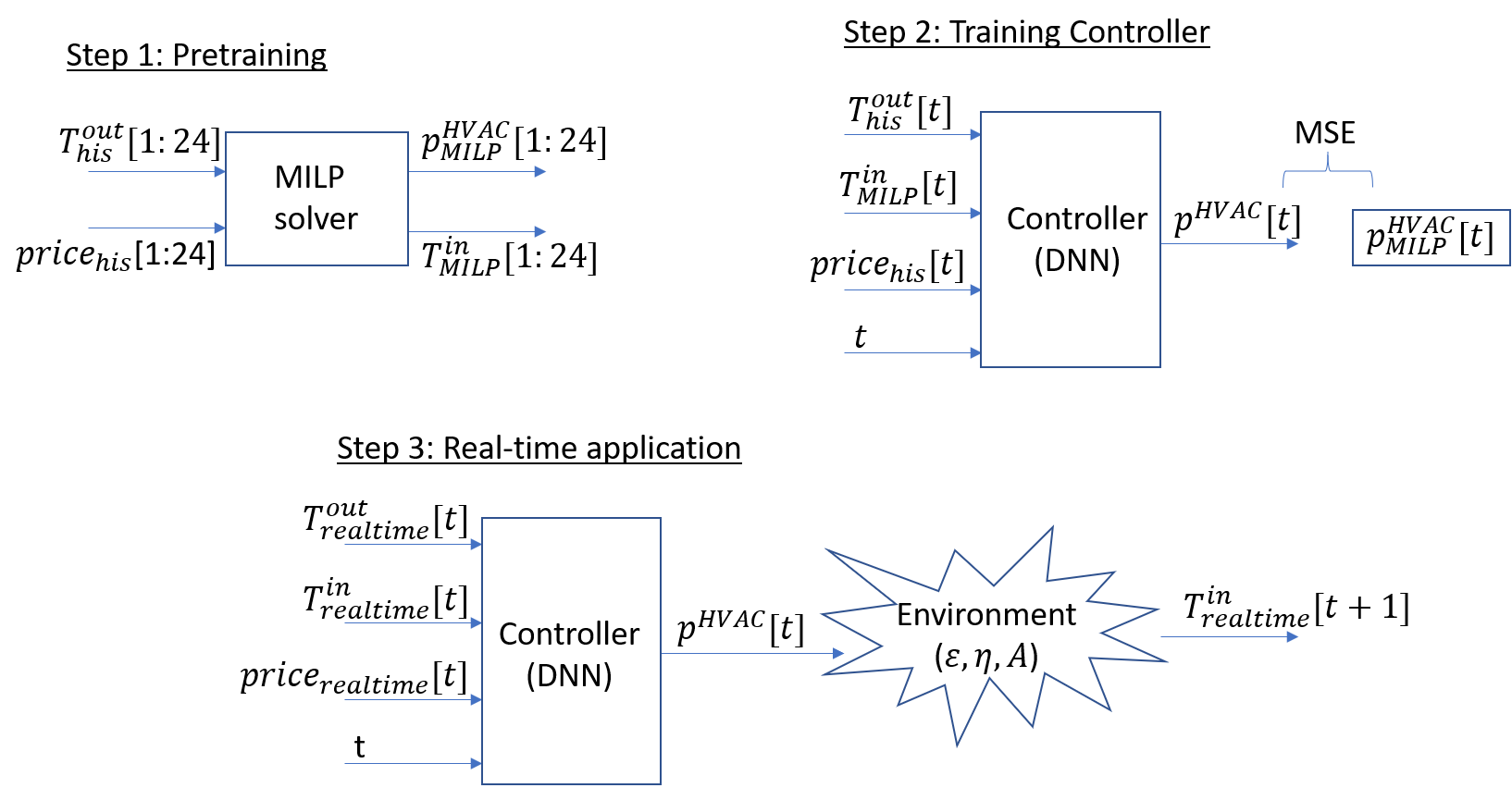}
\caption{Overall framework of MILP-based imitation learning.}
\label{framework}
\end{figure*}

It is clear that our new MO function is a MILP problem. Many MILP solvers can be used to solve our function, and they give us the optimal power consumption the HVAC system should use at each hour in a day. However, the main requirement of using MILP solvers is that we have to know outdoor temperature $T_t^{out}$ at every time slot of that day. This means that MILP solvers are useful only at the end of that day which will be too late to control HVAC system. Hence, to overcome this problem and utilize powerful MILP solvers, we propose two methods in the following subsections.

\subsection{MILP-based Imitation Learning}

The primary goal of imitation learning is to mimic expert behavior in a given task \cite{hussein2017imitation}. First, an agent is trained to learn a general policy by a human or by optimal actions that are already know for each situation in its environment. After training, this agent will use this policy to interact with its environment in the future.

In this method, the HVAC system will mimic the behaviors of a MILP solver. To help the HVAC system interact with the environment the same as MILP behavior in real time, a controller (an agent) is embedded in the HVAC system. This controller will be trained based on optimal actions of the MILP solver using historical outdoor temperatures in the past. After training, the controller applies its knowledge to control HVAC system with real-time data.

Fig. \ref{framework} shows the overall framework of our approach. In step 1 of Fig. \ref{framework}, the historical outdoor temperature, $T_{his}^{out}$, and the historical prices, $price_{his}$, of 24 time slots of a historical day are input data for the MILP solver. The output of the MILP solver is the optimal input powers of HVAC system, $p_{MILP}^{HVAC}$, which should be set up at each time slot of that day to satisfy both objectives: energy cost reduction and thermal comfort. In addition, the MILP solver also gives us optimal indoor temperature, $T_{MILP}^{in}$, at every time slot of that day.

\begin{figure}[ht]
\centering
\includegraphics[scale=0.4]{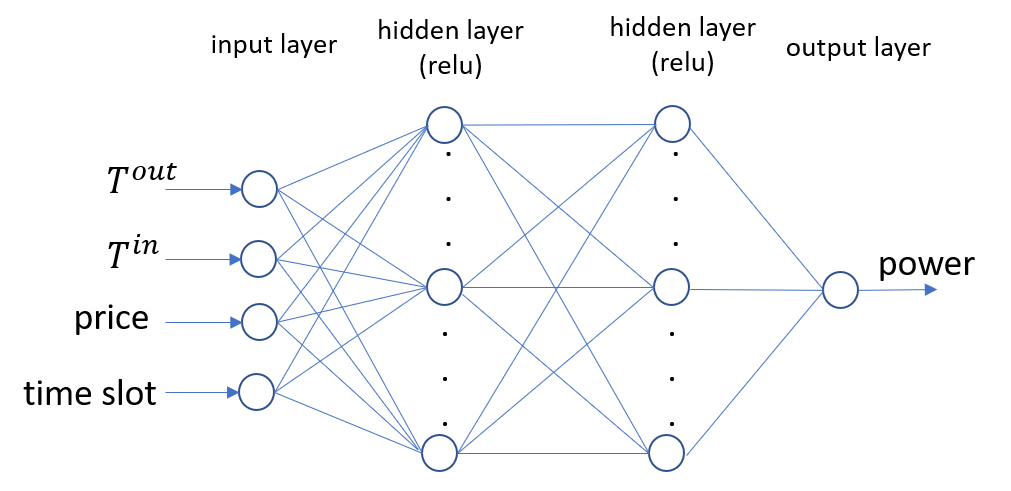}
\caption{Architecture of the controller.}
\label{DNN}
\end{figure}

In step 2, all data created by the MILP solver are used for training the controller of the HVAC system. Our controller is a deep neural network (DNN) whose architecture is shown in Fig. \ref{DNN}. The input data for this DNN consist of four types of information: historical outdoor temperature, $T_{his}^{out}[t]$, at time slot $t$; historical optimal indoor temperature,  $T_{MILP}^{in}[t]$, at time slot $t$, which is calculated by the MILP solver; historical price, $price_{his}[t]$, at time slot $t$; and time slot $t$. The output of this DNN is the input power $p_t$ of the HVAC system during time slot $t$. This output is labeled by the optimal input powers, $p_{MILP}^{HVAC}[t]$, at time slot $t$, which is calculated by the MILP solver in the previous step.

Finally, our trained controller mimics the MILP solver to control the HVAC system in real life, as shown in step 3 of Fig. \ref{framework}. At each time slot in a day, $T_{realtime}^{out}[t], T_{realtime}^{in}[t], price[t]$, and time slot $t$ are input data for our controller (DNN), and its output is the power consumption, $p^{HVAC}[t]$ that the HVAC system should use in this time slot. It is worth noting that, in this step, all input data of our controller are current real-time values of the prices and the environment.

The detailed algorithm of the MILP-based imitation learning method can be found in Algorithm \ref{Real_algorithm}.

\begin{algorithm}
  \caption{MILP-based Imitation Learning method}
  \label{Real_algorithm}
  \begin{algorithmic}[1]
  \State \textit{Step 1: Create training dataset from historical days.}
  \State \textbf{Input:} number of historical days, $hd$, historical outdoor temperature, $T_{his}^{out}[hd][24]$, and historical day-ahead prices, $price_{his}[hd][24]$
    \State \textbf{Output:} A dataset to train the DNN including: array $inputs$ and array $labels$
  \State $inputs=[]$
  \State $labels=[]$
  \For{$i$=1,2,...,$hd$}
    \State /* Finding optimal HVAC power inputs for a historical day by using a MILP solver*/
  	\State Solve (\ref{MOF_2}) with  $T_{his}^{out}[i]$, $price_{his}[i]$ $\implies$ $p_{MILP}^{HVAC}[1:24]$ and $T_{MILP}^{in}[1:24]$
  	\State
  	\For{$t$=1,2,...,24}
  	   \State $inputs.append([T_{his}^{out}[i][t], T_{his}^{in}[i][t], price_{his}[i][t], t])$
  	   \State $labels.append(p_{MILP}^{HVAC}[t])$
  	\EndFor
  	\State
  \EndFor
  \State
  \State \textit{Step 2: Training a DNN, $\theta$, with $inputs$ and $labels$.}
  \State  
  \State \textit{Step 3: Using DNN, $\theta$, in real time with environmental parameters: $\varepsilon, \eta, A$.}
  \For{$t$=1,2,...,24}
     \State measure outdoor temperature $T_{realtime}^{out}[t]$
     \If{$t$==1}
		 \State measure indoor temperature $T_{realtime}^{in}[t]$
	 \Else
		\State calculate indoor temperature $T_{realtime}^{in}[t]$ as in (\ref{formula_input_temperature})			 
	 \EndIf
  	 \State $p = \theta(T_{realtime}^{out}[t], T_{realtime}^{in}[t], price_{realtime}[t], t)$
  	 \State
  	 \State $p^{HVAC}[t] = p$
  	 \State Control HVAC sytem based on power consumption $p^{HVAC}[t]$ for 1h 
    \EndFor
  \end{algorithmic}
\end{algorithm}

\subsection{Forecast-based MILP method}
\begin{figure}[ht]
\centering
\includegraphics[scale=0.4]{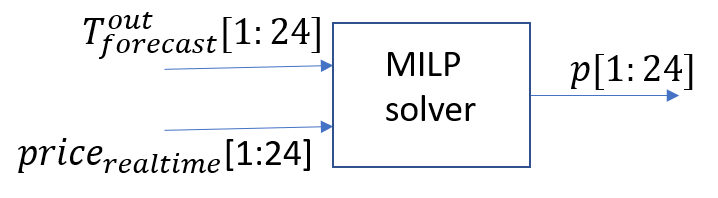}
\caption{Framework of forecast-based MILP method.}
\label{forecast}
\end{figure}

The MILP solver always gives us optimal results (the power consumption at every time slot in a day). However, the main prerequisite of using the MILP solver is that we have to know all the information of that day, and we usually do not have this information until the end of that day. We can overcome this problem by using forecast information that can be collected from some sources such as the National Weather Service (NWS) \cite{NWS}. Fig. \ref{forecast} shows the framework of this method.

As shown in Fig. \ref{forecast}, 24-hour-forecast outdoor temperatures and day-ahead prices are input data for the MILP solver. The output of this solver is the power consumption at each time slot of the next day which is used to control the HVAC system in that day. It is clear that the results of this method depend on accuracy of the forecast information. The accuracy of temperature forecasts depends on many factors such as the season, location and the weather. For example, forecast error is smallest in summer and largest in winter \cite{martner1999five}. 

\section{Simulations and Discussions}
In this section, we evaluate the performance of our proposed methods. We first describe the simulation setup. Next, we compare our methods with the optimal results calculated at the end of day by the MILP solver using the real outdoor temperatures and the real day-ahead prices.

\subsection{Simulation Setup}
In this study, a HVAC system installed in a small home in Detroit city, Michigan, United States is used in the simulation. For simplicity, the heating mode of the HVAC system is considered. Main parameters of this system and the environment are shown in Table \ref{main_parameters}. The DNN of our controller includes four layers, comprising one input layer with 4 neurons, two hidden layers with 100 neurons for each layer, and one output layer with 1 neuron.

\begin{table}[ht]
\caption{Main parameters of HVAC system and environment.}
\centering
\begin{tabular}{|c|c|}
\hline
\textbf{Parameters}	& \textbf{Value}\\
\hline
$T_{min}$ &  $66.2$ $^\text{o}$F ($19$ $^\text{o}$C)\\
\hline
$T_{max}$ &  $75.2$ $^\text{o}$F ($24^\text{o}$C)\\
\hline
$P_{max}$ &  $15$ kW\\
\hline
$\varepsilon$ &  $0.7$ \cite{deng2016indoor}\\
\hline
$\eta$ &  $2.5$ \cite{yu2019deep}\\
\hline
$A$ &  $0.14$ kW/$^\text{o}$F \cite{yu2019deep}\\
\hline
$ DNN $ &  $4:100:100:1$\\
\hline
$ Optimizer $ &  $Adam$\\
\hline

\end{tabular}
\label{main_parameters}
\end{table} 

\begin{figure}[ht]
\centering
\includegraphics[scale=0.3]{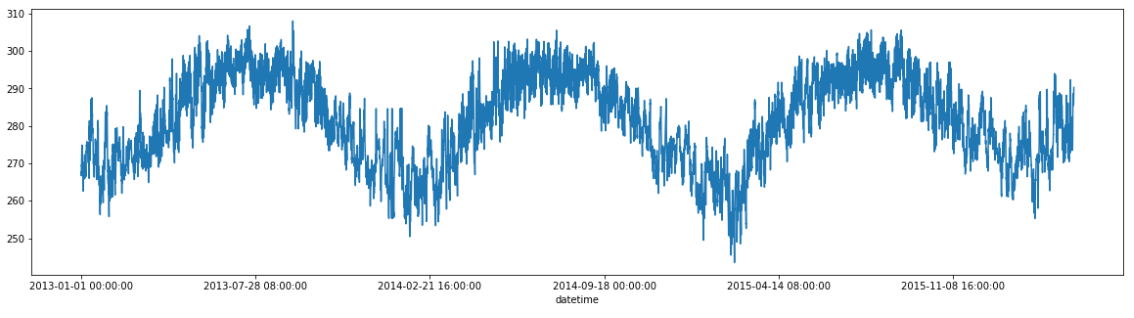}
\caption{Outdoor temperatures of Detroit city from 2013 to 2016.}
\label{outdoor_temperature}
\end{figure}

\begin{figure}[ht]
\centering
\includegraphics[scale=0.3]{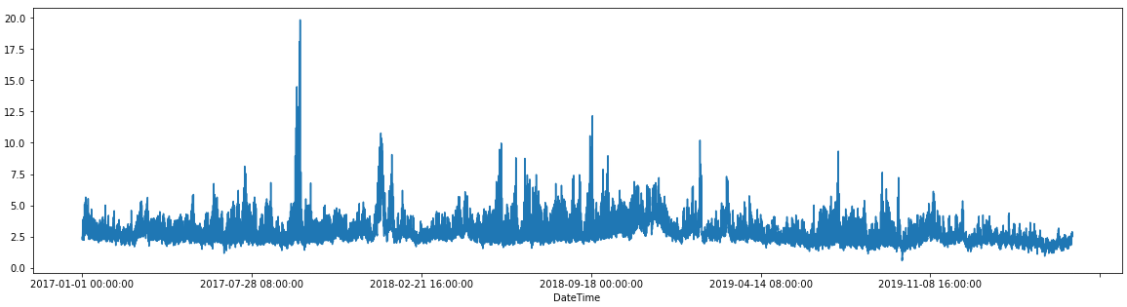}
\caption{Day-ahead prices in Michigan state from 2017 to 2020.}
\label{DAP}
\end{figure}

For training and testing our controller, the real outdoor temperatures of Detroit city from 2013 to 2016 are obtained from the Kaggle website \cite{kaggle}, as shown in Fig. \ref{outdoor_temperature}. Because we do not have day-ahead prices for these years, day-ahead prices from 2017 to 2020 obtained from Pecan Street database\footnote{https://www.pecanstreet.org} for Michigan state are used as shown in Fig. \ref{DAP}. To be specific, the real outdoor temperatures from January 1, 2013, to December 31, 2015, and day-ahead prices from January 1, 2017, to December 31, 2019, are used to train our controller. Our controller is then run for each day from January 1, 2016, to March 31, 2016, with day-ahead prices from January 1, 2020, to March 31, 2020, to test its performance. During training and testing, we assume that at the beginning of a day, with no heating sources running, the indoor temperature at time slot $1$ is equal to the outdoor temperature. It is noteworthy that in all the above years, summer data are removed because the outdoor temperature of Detroit city is usually higher than $T_{min}$ in the summer season, and only the heating mode of the HVAC system is considered.

\begin{figure}[ht]
\centering
\includegraphics[scale=0.5]{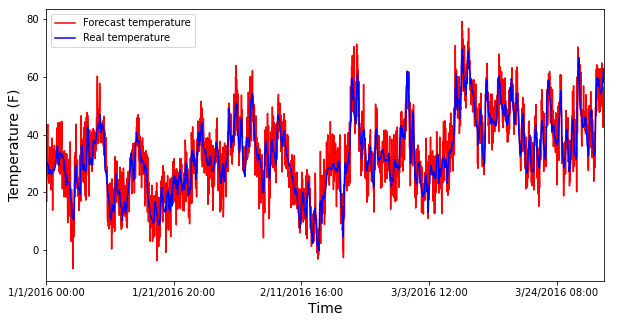}
\caption{Real outdoor temperatures and forecast outdoor temperatures of Detroit city.}
\label{real}
\end{figure}

For comparison, we also require 24-hour-forecast outdoor temperatures of Detroit city from January 1, 2016, to March 31, 2016, and day-ahead prices from January 1, 2020, to March 31, 2020, as input data in forecast-based MILP method. However, we have only the real outdoor temperatures of Detroit city for that time. Some studies have shown that the forecast error of the day-ahead temperature follows certain statistic characteristics \cite{taylor2005single}. Hence, to re-create the forecast temperature of historical days, we add error $E_G$ to the real outdoor temperature at each time slot, as shown in (\ref{error}) where $E_G$ is the Gaussian distribution (normal distribution) with a mean of $0.5^\text{o}F$ and a standard deviation of $6^\text{o}F$ \cite{taylor2005single}, \cite{zhang2016optimal}. The real outdoor temperatures and forecast outdoor temperatures of Detroit city from January 1, 2016, to March 31, 2016, are shown in Fig. \ref{real}.

\begin{equation}
T_{forecast}^{out}[t] = T_{real}^{out}[t] + E_G \hspace{0.5cm} \forall t
\label{error}
\end{equation}

\subsection{The effects of parameter $\alpha$}
Solving (\ref{MOF_2}) using the MILP solver with historical days creates the dataset for training our controller. Table \ref{effects} shows the range of the indoor temperature and the total energy cost for historical days from January  1, 2013, to December 31, 2016, with different values of parameter $\alpha$ in (\ref{MOF_2}).

\begin{table}[ht]
\caption{The effects of parameter $\alpha$.}
\centering
\begin{tabular}{|c|c|c|}
\hline
\textbf{$\alpha$}	& \textbf{Range of indoor temperature ($^\text{o}$C)} & \textbf{Total energy cost (cents)}\\
\hline
$1$ &  9.7-27.7 & 30341.9\\
\hline
$2$ &  18.7-27.7 & 30353.6\\
\hline
$3$ &  18.9-27.7 & 30353.7\\
\hline
$4$ &  19-27.7 & 30353.8 \\
\hline
$5$ &  19-27.7 & 30353.8 \\
\hline

\end{tabular}
\label{effects}
\end{table}

As shown in Table \ref{effects}, with $\alpha = 1$, the range of indoor temperatures is far from the comfortable range, and it is not appropriate for simulation. With $\alpha \geq 2$, the range of indoor temperatures is acceptable and the total energy cost is almost the same. However, $\alpha \geq 4$ gives us the best results, and therefore $\alpha = 4$ is chosen to create the dataset for training our controller and comparing our two methods.

\subsection{Performance of two methods}
First, the performance of the two methods in a day is compared with the optimal results calculated by using the MILP solver at the end of the day after we have all information about the outdoor temperature at every time slot under two metrics: mean absolute error (MAE) and mean absolute percentage error (MAPE) as shown in (\ref{MAE}) and (\ref{MAPE}).

\begin{equation}
MAE = \frac{1}{N} \displaystyle \sum_{n=1}^{N}|V_{method}(n) - V_{MILP}(n)|
\label{MAE}
\end{equation}

\begin{equation}
MAPE = \frac{100}{N} \displaystyle \sum_{n=1}^{N}\frac{|V_{method}(n) - V_{MILP}(n)|}{V_{MILP}(n)} 
\label{MAPE}
\end{equation}
where $N$ is the number of samples, $V_{method}(n)$ denotes the value of our methods, and $V_{MILP}(n)$ refers to the value of the optimal results.

Second, the performance of the two methods is compared based on the values of these metrics under three terms: the hourly power consumption, the daily energy cost and the range of indoor temperatures.

\subsubsection{MAE and MAPE of hourly power consumption and daily energy cost}

\begin{figure}[ht]
\centering
\includegraphics[scale=1]{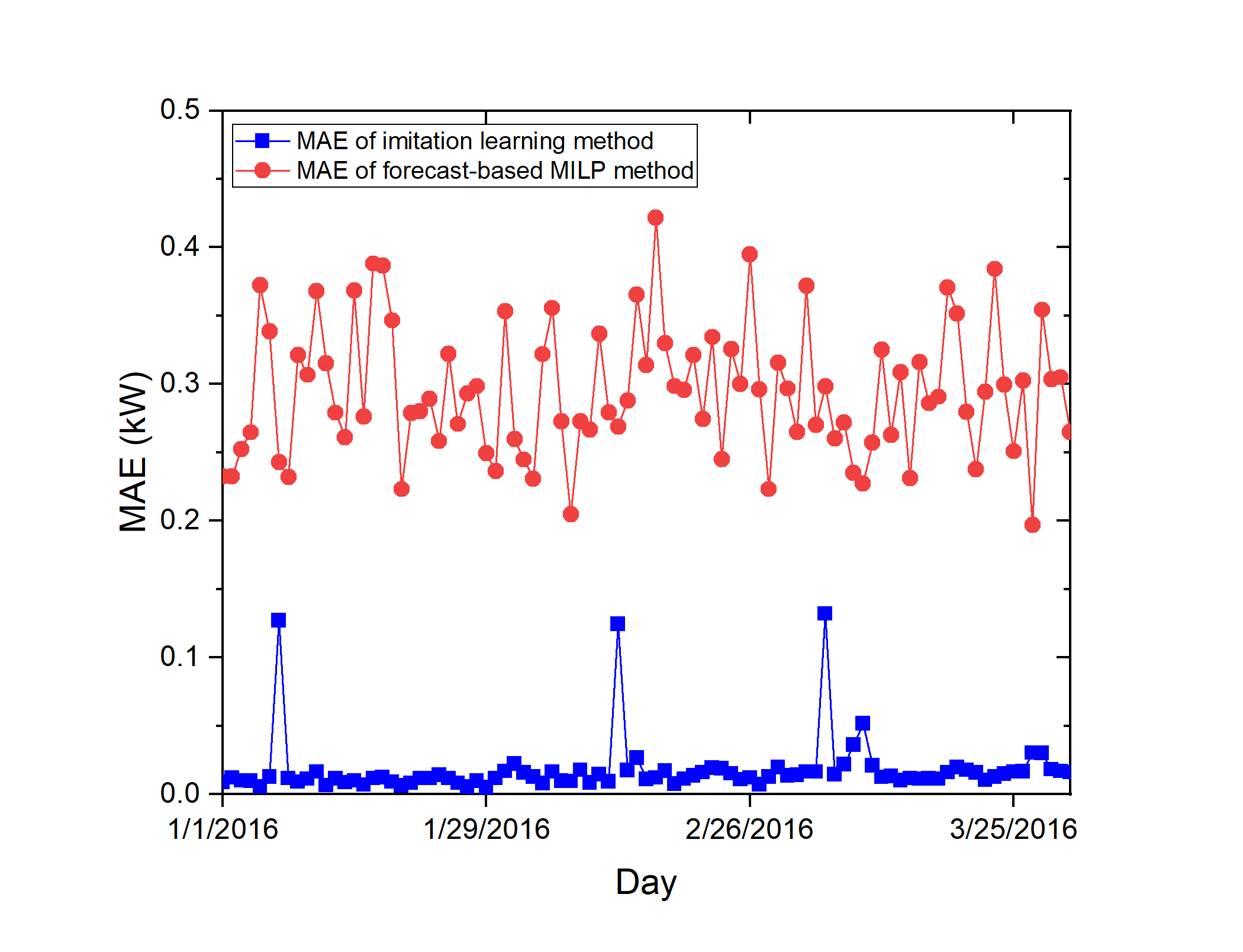}
\caption{MAE of two methods compared with optimal hourly power consumption in a day.}
\label{MAE_power}
\end{figure}

\begin{figure}[ht]
\centering
\includegraphics[scale=1]{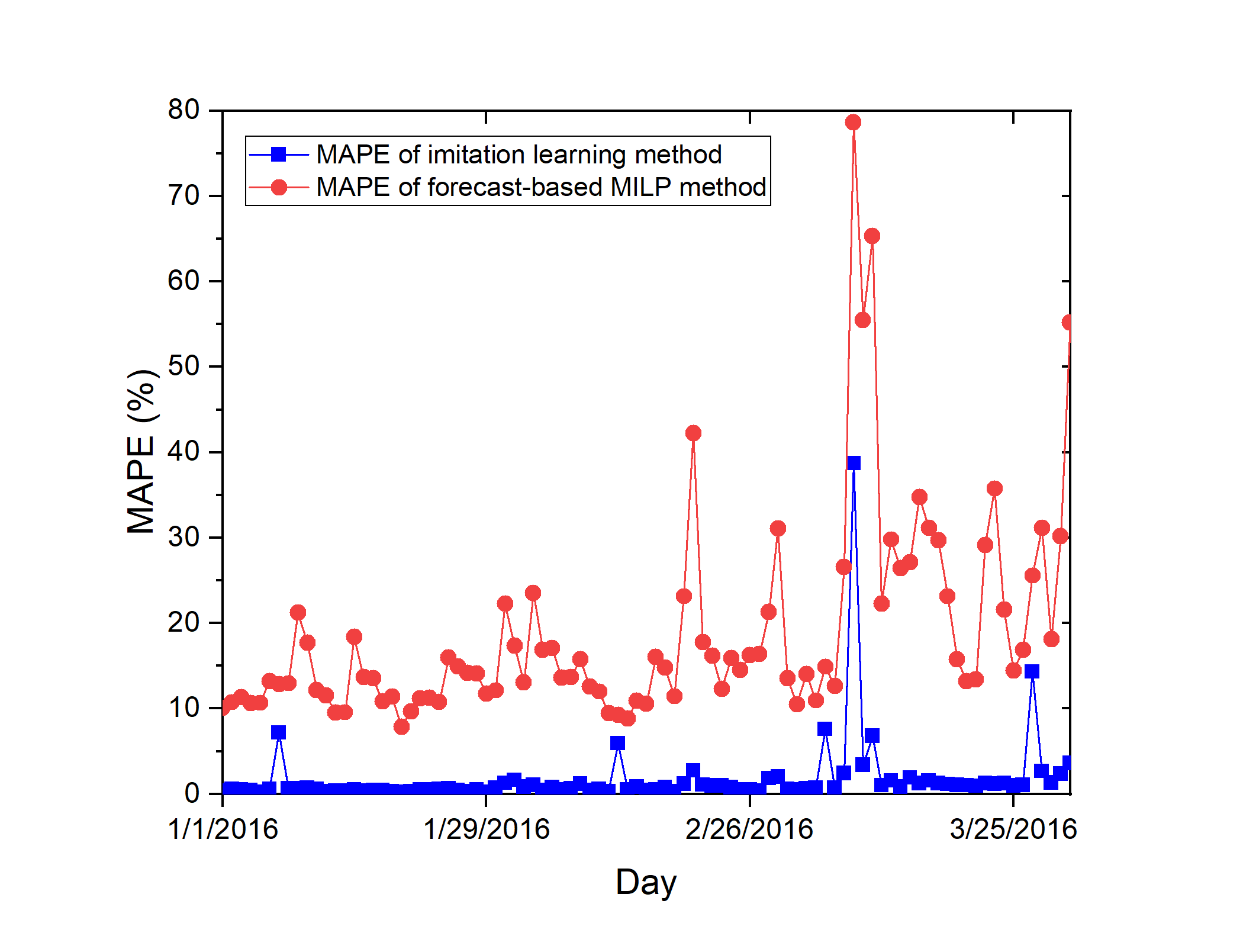}
\caption{MAPE of two methods compared with optimal hourly power consumption in a day.}
\label{MAPE_power}
\end{figure}

Figs. \ref{MAE_power} and \ref{MAPE_power} show the MAE and MAPE of the hourly power consumption of the imitation learning method and the forecast-based MILP method with the hourly optimal power consumption created by the MILP solver at every hour of a day during the winter season of 2016 from January 1, 2016, to March 31, 2016. As depicted in these figures, on most of these days, the difference between the results of the imitation learning method and the optimal results is very small at each hour. The average MAE and MAPE of the hourly power consumption in the imitation learning method in the winter season are only approximately $0.02$ kW and $1.7\%$, respectively. Meanwhile, the difference between the results of the forecast-based MILP method with optimal results is very large. The average MAE and MAPE of the hourly power consumption in the forecast-based MILP method increase to $0.3$ kW and $18.9\%$, respectively. This is because the imitation learning method approximates the MILP solver very well, and the forecast error worsens the performance of the forecast-based MILP method.

\begin{figure}[ht]
\centering
\includegraphics[scale=1]{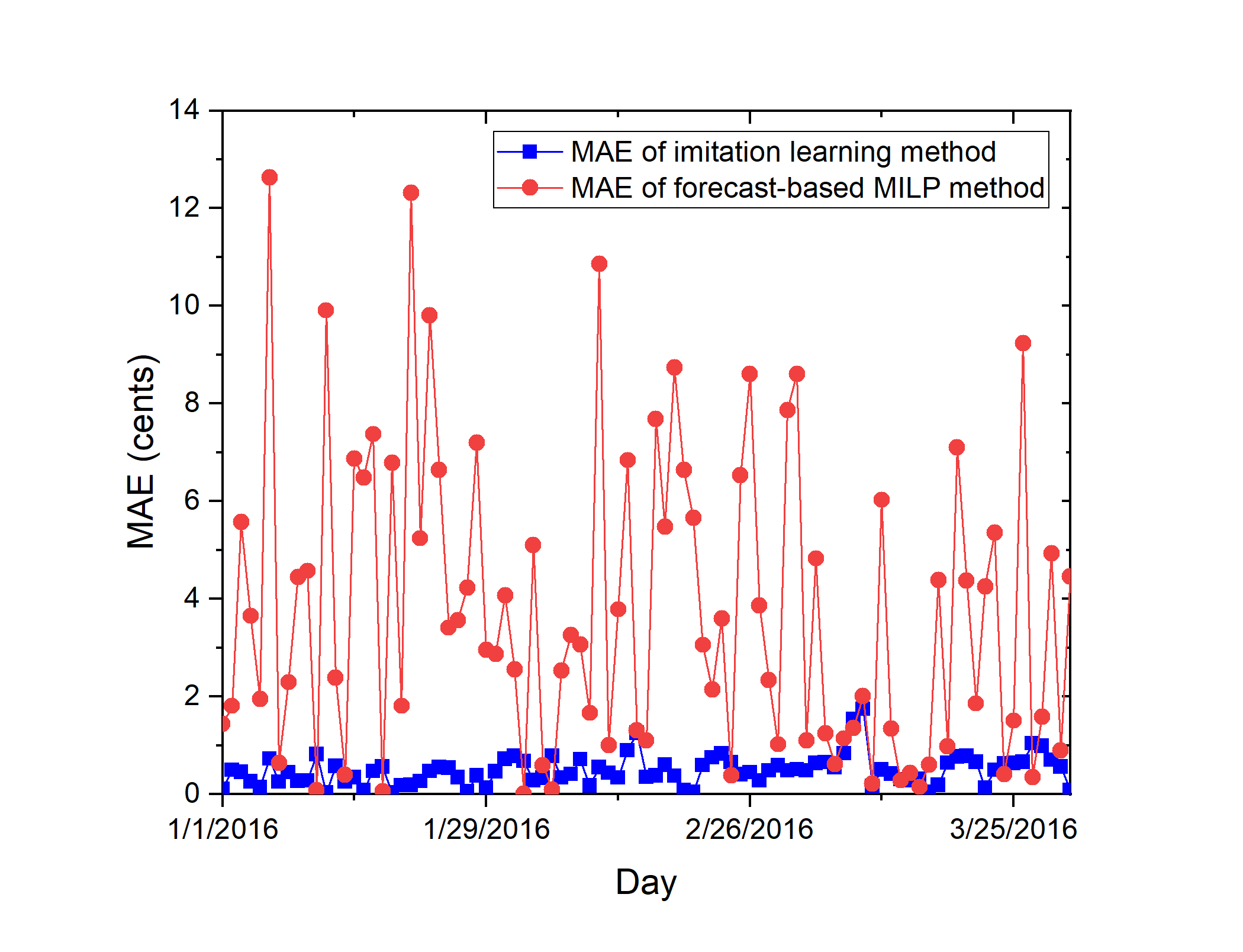}
\caption{MAE of two methods compared with optimal daily energy cost in a day.}
\label{MAE_cost}
\end{figure}

\begin{figure}[ht]
\centering
\includegraphics[scale=1]{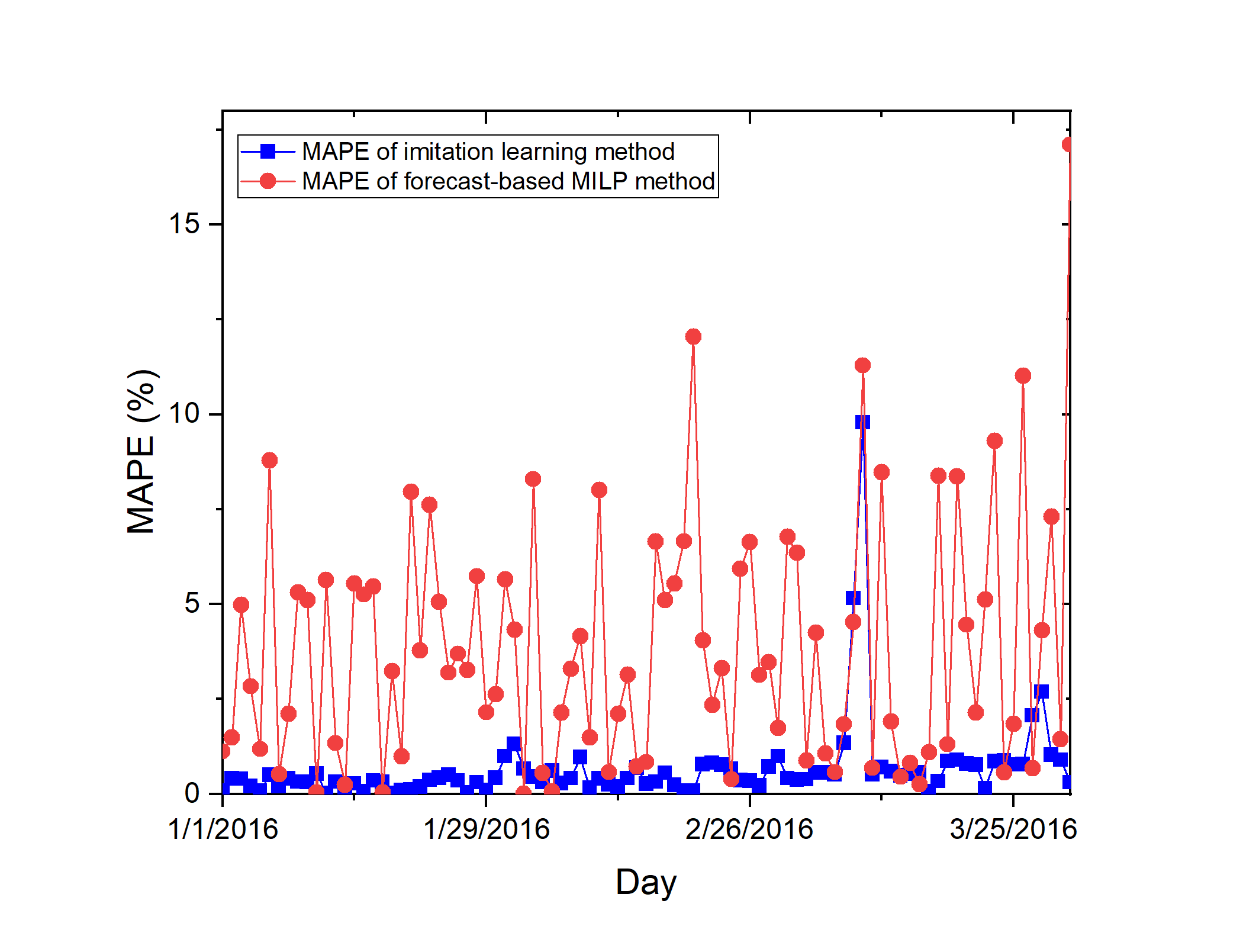}
\caption{MAPE of two methods compared with optimal daily energy cost in a day.}
\label{MAPE_cost}
\end{figure}

For this reason, the performance of the forecast-based MILP method is also worse than the imitation learning method in terms of the daily energy cost, as shown in Figs. \ref{MAE_cost} and \ref{MAPE_cost}. The average MAE and MAPE of the daily energy cost in the forecast-based MILP method during the winter season are respectively approximately $3.8$ cents and $3.9\%$ whereas the average MAE and MAPE of the daily energy cost in the imitation learning method are only approximately $0.5$ cents and $0.7\%$.

A significant disadvantage of the MAPE function is that it produces an infinite or an undefined large value when the $V_{MILP}(n)$ in (\ref{MAPE}) is zero or close to zero. Hence, as shown in Figs. \ref{MAPE_power} and \ref{MAPE_cost}, the MAPEs of some days are very large, compared with those of other days. In these days, the hourly power consumption or daily energy cost is very small. Even though the MAPEs of these days are large, this does not affect the indoor temperatures.

\subsubsection{Range of hourly indoor temperatures}
\begin{figure}[ht]
\centering
\includegraphics[scale=1]{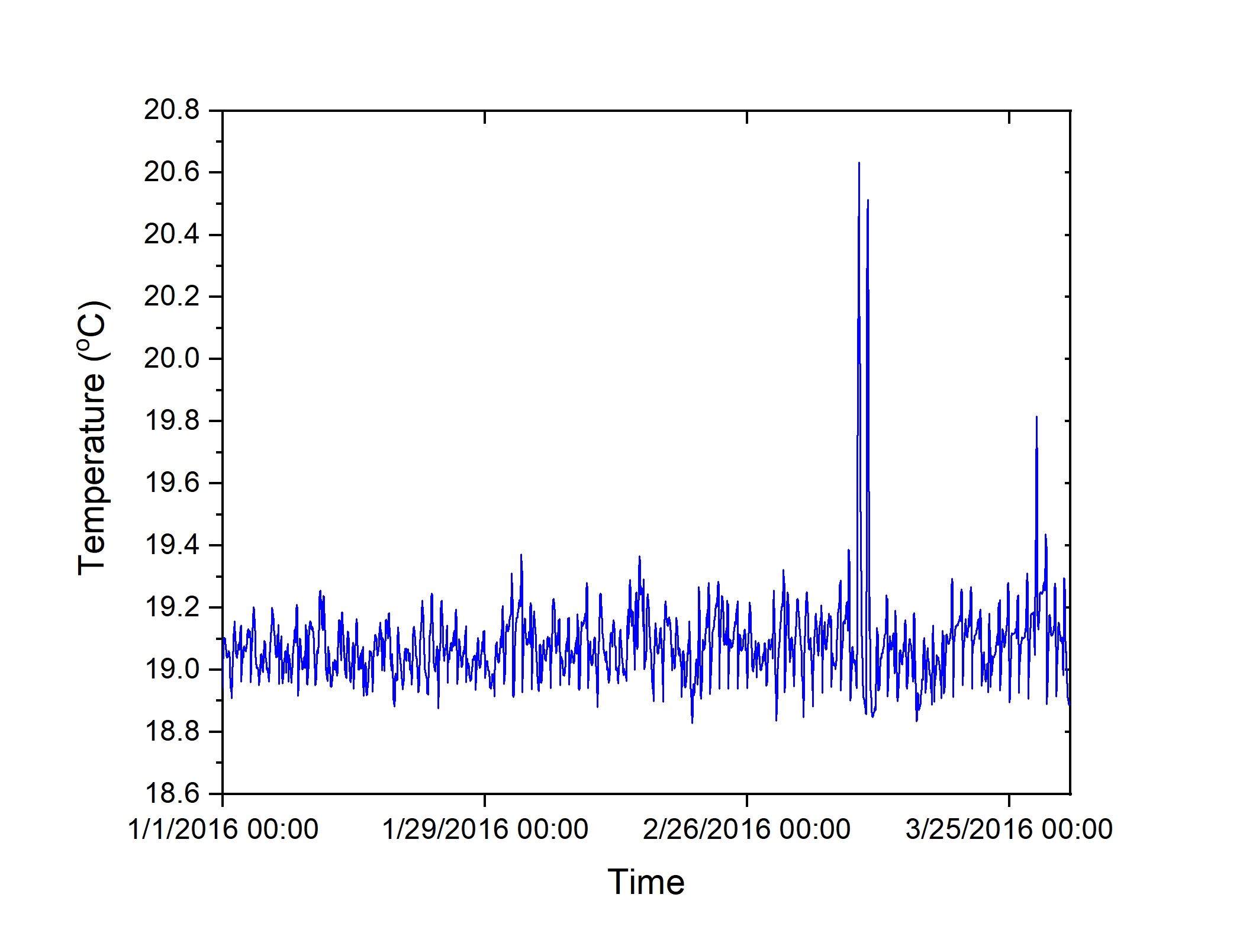}
\caption{Hourly indoor temperatures using the imitation learning method.}
\label{temperature_il}
\end{figure}

\begin{figure}[ht]
\centering
\includegraphics[scale=1]{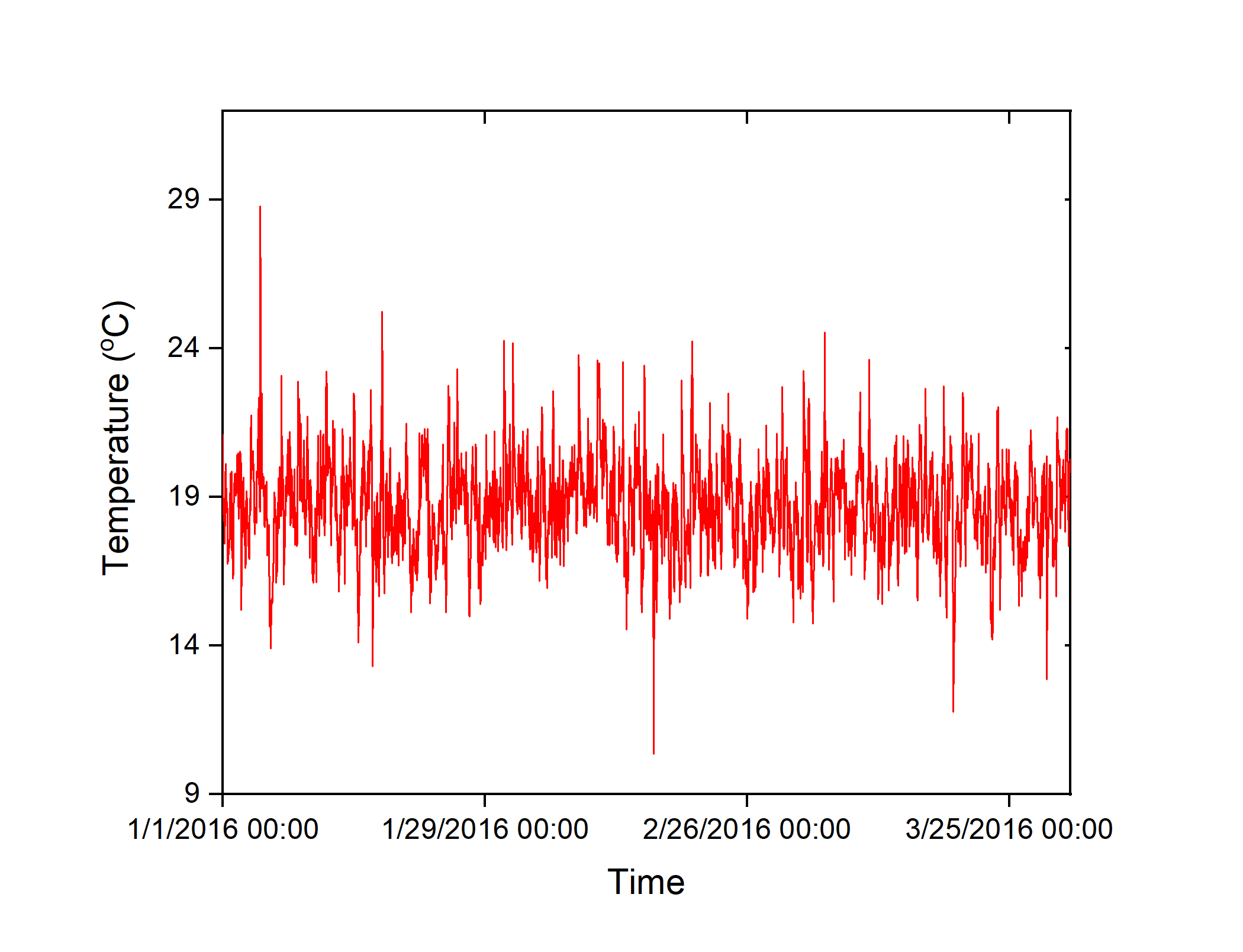}
\caption{Hourly indoor temperatures using forecast-based MILP method.}
\label{temperature_fc}
\end{figure}

One of the main purposes of the HVAC system is to guarantee that the indoor temperature is in a comfortable range from $19$ $^\text{o}$C to $24$ $^\text{o}$C. As shown in Fig. \ref{temperature_il}, during the winter season of 2016, the hourly indoor temperature set in the imitation learning method ranges from $18.8$ $^\text{o}$C to $20.6$ $^\text{o}$C, and its mean value is $19.1$ $^\text{o}$C. In the meanwhile, the hourly indoor temperature set in the forecast-based MILP method ranges from $10.3$ $^\text{o}$C to $28.8$ $^\text{o}$C, and its mean value is $18.7$ $^\text{o}$C, as shown in Fig. \ref{temperature_fc}. It is clear that the range of indoor temperatures in the imitation learning method is very close to the comfortable range, whereas this range in the forecast-based MILP method is far from the comfortable range at both the upper and lower bounds. Because the mean values of both methods are similar and very close to $19$ $^\text{o}$C, we use the standard deviation to compare the performance of the two methods, as shown in (\ref{deviation}).

\begin{equation}
\sigma = \sqrt{\frac{(T^{in}[n]-\mu)^2}{N}}
\label{deviation}
\end{equation}
where $\mu$ is the mean value, $N$ is the number of samples

As shown in Table \ref{dt}, the deviation in the temperature set in the imitation learning method is $0.23$, whereas this value in the forecast-based MILP method is $2.91$. These values show that the indoor temperatures in the forecast-based MILP method spread farther from the mean value than the in imitation learning method. As shown in Fig. \ref{temperature_fc}, there are many hours at which the indoor temperature is very low or very high. The performance of the imitation learning method is better than that of the forecast-based MILP method for satisfying thermal comfort.

\begin{table}[ht]
\caption{The deviation in indoor temperature set of two methods.}
\centering
\begin{tabular}{|c|c|}
\hline
\textbf{Method}	& \textbf{Deviation} \\
\hline
Imitation learning method & 0.23\\
\hline
Forecast-based MILP method &  2.91\\
\hline

\end{tabular}
\label{dt}
\end{table}

\section{Conclusions and future works}
In this paper, we proposed two methods for controlling a HVAC system to provide thermal comfort and reduce the energy cost in a day. In the first method, a controller (DNN) is used as an imitator to mimic the optimal results of the MILP solver which uses historical data; its knowledge is then applied to current real-time data. In the second method, 24-hour-forecast temperatures are used as input data for the MILP solver, and the HVAC system is controlled based on the output data of the MILP solver.

Numerical results show that the performance of the imitation learning method is better than that of the forecast-based MILP method in terms of hourly power consumption, daily energy cost, and thermal comfort. These results also show that the imitation learning method produces results that are very similar to the optimal results, which we can obtain only from the MILP solver at the end of a day. The differences between these results are very small, only $0.02$ kW and $1.7\%$ for the average MAE and MAPE in terms of the hourly power consumption and $0.5$ cents and $0.7\%$ for the average MAE and MAPE in terms of the daily energy cost. Moreover, the range of indoor temperatures in the imitation learning method is very close to the comfortable range. These results prove that our controller successfully approximates a MILP solver.

However, it should be mentioned that our study still has some limitations in practice. First, we are interested in integrating many factors in building thermal comfort, such as humidity and solar radiation. Second, extending our imitation learning method for energy management is another way to improve our system. The future imitation learning method will include many DNNs, and each DNN will control an electronic device or a space (zones) in smart homes.


%





\ifCLASSOPTIONcaptionsoff
  \newpage
\fi



\bibliographystyle{IEEEtran}
\bibliography{IEEEabrv,my_references}
\end{document}